\documentclass[10pt,twocolumn,letterpaper]{article}

\usepackage[pagenumbers]{cvpr} 

\usepackage{graphicx}
\usepackage{amsmath}
\usepackage{amssymb}
\usepackage{booktabs}
\usepackage{adjustbox}
\usepackage[dvipsnames]{xcolor}
\usepackage{colortbl}
\usepackage{float}
\usepackage{lipsum}  
\usepackage{multirow}
\usepackage{microtype}
\usepackage[accsupp]{axessibility}
\usepackage{enumitem}

\usepackage{hyperref}
\hypersetup{colorlinks}

\usepackage[capitalize]{cleveref}
\crefname{section}{Sec.}{Secs.}
\Crefname{section}{Section}{Sections}
\Crefname{table}{Table}{Tables}
\crefname{table}{Tab.}{Tabs.}

\def\cvprPaperID{8638} 
\def\confName{CVPR}
\def\confYear{2023}

\begin{document}
\newcommand{\stda}{\textit{SelTDA}}
\newcommand{\authsep}{  }

\title{Q: How to Specialize Large Vision-Language Models to Data-Scarce VQA Tasks?\\ A: Self-Train on Unlabeled Images!}

\author{\normalsize{Zaid Khan$^{\dagger\ast}$ \authsep Vijay Kumar BG$^{\clubsuit}$ \authsep Samuel Schulter$^{\clubsuit}$ \authsep Xiang Yu$^{\diamondsuit\ast}$ \authsep Yun Fu$^{\dagger}$ \authsep Manmohan Chandraker$^{\clubsuit\heartsuit}$}\\
$^{\dagger}$Northeastern University, $^{\clubsuit}$NEC Labs America, $^{\diamondsuit}$Amazon, $^{\heartsuit}$UC San Diego 
}
\maketitle

{\let\thefootnote\relax\footnote{
{$^{\ast}$work done while at NEC Labs America}}}

\begin{abstract}

Finetuning a large vision language model (VLM) on a target dataset after large scale pretraining is a dominant paradigm in visual question answering (VQA).
Datasets for specialized tasks such as knowledge-based VQA or VQA in non natural-image domains are orders of magnitude smaller than those for general-purpose VQA.
While collecting additional labels for specialized tasks or domains can be challenging, unlabeled images are often available.
We introduce \stda~ (\textbf{Sel}f-\textbf{T}aught \textbf{D}ata \textbf{A}ugmentation), a strategy for finetuning large VLMs on small-scale VQA datasets.
\stda~uses the VLM and target dataset to build a teacher model that can generate question-answer pseudolabels directly conditioned on an image alone, allowing us to pseudolabel unlabeled images.   
\stda~then finetunes the initial VLM on the original dataset augmented with freshly pseudolabeled images.
We describe a series of experiments showing that our self-taught data augmentation increases robustness to adversarially searched questions, counterfactual examples and rephrasings, improves domain generalization, and results in greater retention of numerical reasoning skills.
The proposed strategy requires no additional annotations or architectural modifications, and is compatible with any modern encoder-decoder multimodal transformer.
Code available at \url{https://github.com/codezakh/SelTDA}.
\end{abstract}

\section{Introduction}
\begin{figure}
    \centering
    \includegraphics{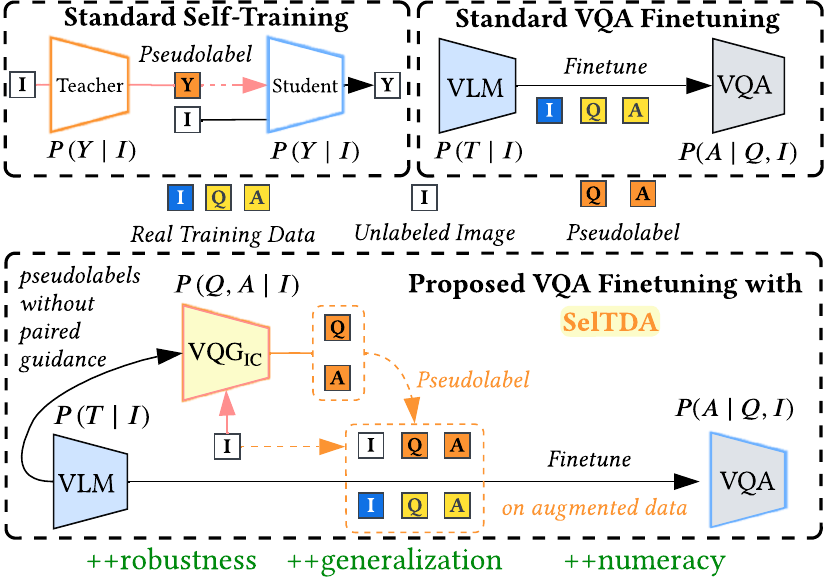}
    \caption{\stda~expands the self-training paradigm to VQA. 
    By self-generating supervision (\textcolor{orange}{orange line}) for an image \textit{I} without needing extra annotations, we can augment a target dataset with new images and their pseudo-questions and answers $(Q, A)$.  } 
    \label{fig:teaser}
\end{figure}

\begin{figure*}
    \centering
    \includegraphics{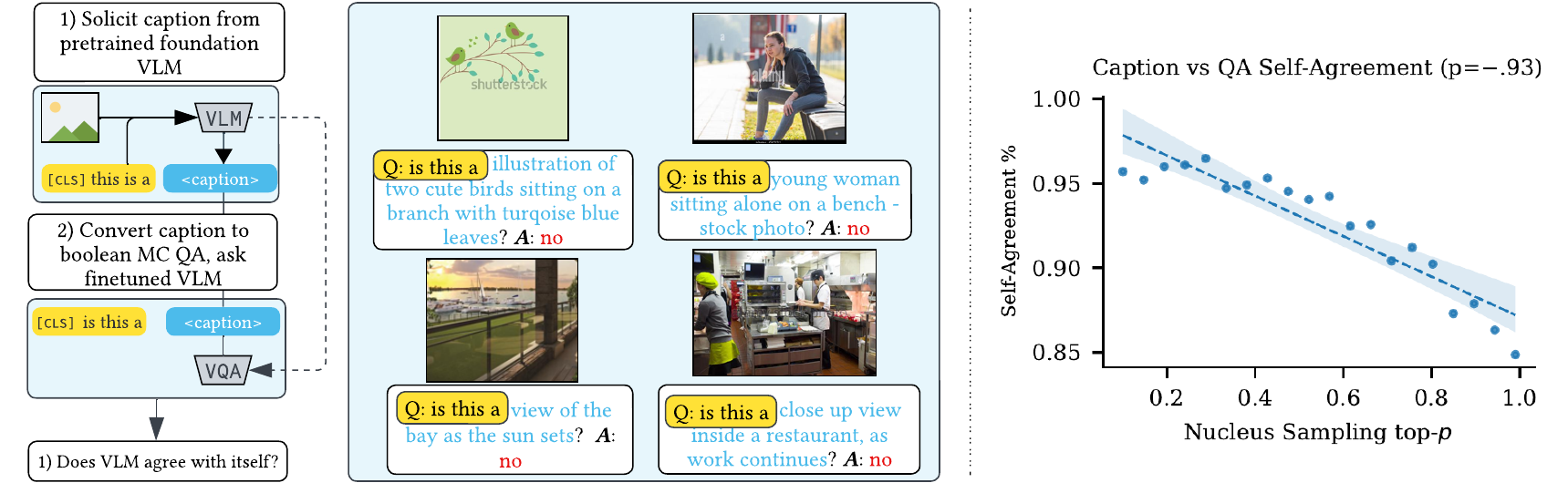}
    \caption{Motivating experiment. We sample increasingly diverse captions from BLIP \cite{blip}, convert them to questions, and pose the questions to BLIP after finetuning on VQAv2. As caption diversity increases, self-agreement decreases (right panel). Despite the diversity, many captions remain correct (middle panel), suggesting that the VLM has knowledge that is not exhausted by task-specific finetuning.}
    \label{fig:dark-knowledge}
\end{figure*}

Large, pretrained vision language foundation models \cite{Bommasani2021OnTO,blip_2, SIMLA, blip, unified_io,ofa} are approaching human-level performance on visual question answering (VQA) \cite{blip,wang2022ofa,Wang2022ImageAA,Wang2021VLMoUV,Yuan2021FlorenceAN,Wang2022SimVLMSV}, as measured by the standard VQAv2 \cite{vqav2} benchmark.
Yet on more complex VQA tasks \cite{AOKVQA,Marino2019OKVQAAV} there is a larger gap between humans and machines. 
One difficulty is the small scale of datasets for complex VQA tasks or those in domains beyond natural images. 
The first solution to deal with the data scarcity is to employ transfer learning from a larger VQA dataset (e.g. VQAv2) to the smaller, specialized VQA dataset. 
However weaknesses of VQA models such as lack of consistency \cite{vqa_rephrasings}, weakness to adversarially searched questions \cite{advqa} and tendency to cheat by learning shortcuts \cite{vqa_ce} can be exacerbated when fine-tuning on small datasets.

Collecting annotations to expand a dataset for knowledge-intensive tasks or specialized domains is often prohibitively expensive.
However, \textit{unlabeled images} are cheap and often available. 
How can we exploit unlabeled images for specific visual question answering tasks? One possibility is to generate new question+answer pairs for the unlabeled images, and use them during training. 
However, existing methods for visual question \textit{generation} require images with annotations --- either ground truth captions \cite{Banerjee2021WeaQAWS,Changpinyo2022AllYM}, or bounding boxes \cite{Kil2021DiscoveringTU,Vedd2022GuidingVQ}.
Even if these annotations were to be acquired, they induce a limited set of possible questions; they are limited to objects and concepts included in the acquired annotation, which are in turn limited by the finite label space of pretrained object detectors and the information disparity between a caption and an image (an image usually contains much more content than a short caption can describe).

\noindent \textbf{Motivating Experiment}: In Fig \ref{fig:dark-knowledge}, we show that a large vision-language model (VLM) pretrained on web-scale data contains knowledge that can be drawn out with image-conditional text generation, but which the model cannot verify when posed as a visual question-answering task. 
We prompt the BLIP \cite{blip} VLM (pretrained on 129M image-text pairs) to caption 1000 images from the CC3M \cite{cc3m} dataset starting with the phrase ``\texttt{this is a}''.
We convert each caption into a boolean question where the correct answer is ``yes'' by inserting the caption into the template \texttt{is this a <caption>?} Next, we ask a BLIP VLM finetuned on the VQAv2 dataset \cite{vqav2} to choose between ``yes'' and ``no'' for each caption turned into a question.
Surprisingly, the VQA-finetuned BLIP answers ``no'' to \textit{at least} $5\%$ of the questions, increasing to $15$\% as the diversity of captions increases (adjusted by top-$p$ parameter in nucleus sampling).
This suggests the possibility that the VLM has knowledge it cannot exploit when answering questions, but is accessible when directly generating text conditioned on an image.

\noindent \textbf{Approach}: To exploit unlabeled images for VQA, we propose \stda, a three-stage framework for \textbf{Sel}f-\textbf{T}aught \textbf{D}ata \textbf{A}ugmentation (Fig \ref{fig:teaser} bottom panel).
We adapt the paradigm of self-training used in object detection \cite{DBLP:conf/nips/ZophGLCLC020,Li2020ImprovingOD} and image classification \cite{4129456,9577846} for VQA. 
In classification / detection, the task of labeling an image is identical to prediction, and the teacher and student optimize identically structured objectives.
In VQA self-training, the student and teacher tasks are different. A teacher must pose and answer a question given an image, while the student provides an answers given a question and image. 
To handle this, we first cast the task of the teacher as a direct image-to-text generation task, and introduce a teacher model by updating the weights of the VLM to learn an \textit{image-conditional} visual question generation model VQG$_{IC}$.
Next, we use VQG$_{IC}$ as a teacher to pseudolabel unlabeled images by sampling questions and answers from VQG$_{IC}$ with stochastic decoding.
Finally, we augment the original VQA dataset with the newly labeled image-question-answer pairs, and finetune the VLM for visual question answering on the augmented VQA dataset.

\noindent \textbf{Benefits}: \stda~ allows us generate synthetic training data by approximating the distribution $P(Q,A|I)$ of the target VQA task, where $Q,A,I$ represents a question, answer, and image respectively.
One benefit is that the synthetic data increases the number of training pairs available for finetuning, which effects an increase in raw performance.
A second benefit is an increase in the diversity of questions and answers due to the introduction of new images and the stochastic nature of the text decoding, which results in increased robustness and domain generalization.
A third benefit is the distillation of knowledge from pretraining and transfer learning into the synthetic training data, which can teach new skills (e.g. domain generalization) or prevent the forgetting of specific skills (e.g. numerical reasoning).
Finally \stda~ is architecture-agnostic given a vision-language model capable of image-conditional text-generation.
Our contributions can be summarized as follows:
\begin{enumerate}
    \item We introduce \stda, a variant of the self-training paradigm that is designed for VQA and large generative pretrained VLMs.
    \item We propose treating visual question generation as a direct image-to-text task by leveraging the autoregressive decoder of a large, pretrained VLM, enabling us to generate questions and answers from an unlabeled image with no auxillary annotations needed.
    \item We show that a large VLM trained with the proposed \stda~gains increased robustness, domain generalization, numerical reasoning, and performance when finetuning on small-scale VQA datasets.
\end{enumerate}

\begin{figure*}
    \centering
    \includegraphics{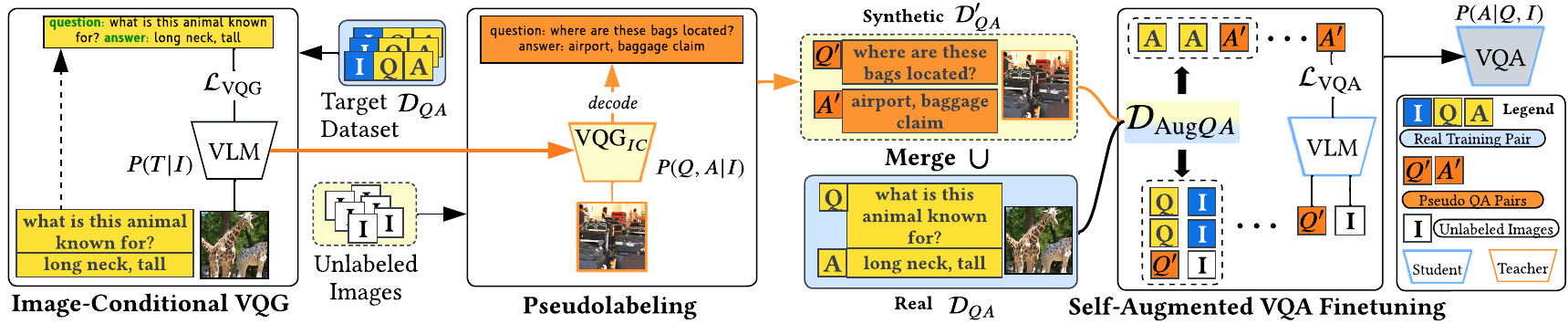}
    \caption{Overview of the proposed framework. We first create the teacher VQG$_{IC}$ (\S\ref{sec:teacher}), use VQG$_{IC}$ to pseudolabel unlabeled images (\S\ref{sec:pseudolabeling}), and finetune student on the original training pairs augmented with the pseudolabeled images. The pseudolabels are natural language.}
    \label{fig:arch-diag}
\end{figure*}
\section{Related Work}
\textbf{Augmentation for VQA} The method of \cite{Wang2021CrossModalGA} augments images by using an MLP to classify possible answers in the image and using an LSTM to generate questions matching the answer. 
While this works with unlabeled images, it is not used for self-training, has a limited label space, and does not leverage large VLMs.
KDDAug\cite{Chen2022RethinkingDA} augments existing question answer pairs by generating pseudoanswers and achieves increases in robustness.
ConCat \cite{Kant2020ContrastAC} similarly trains more robust models by augmenting the \textit{existing} QA pairs in a dataset.
In contrast to this line of work, we seek to exploit \textit{unlabeled images} by generating \textit{new} questions and answers, and using a large VLM to generate augmentation.

\textbf{Few/Zero-shot Generalization} Large VLMs have shown impressive generalization to unseen tasks after large-scale pretraining \cite{Alayrac2022FlamingoAV}, echoing similar achievements in natural language processing \cite{Wei2022FinetunedLM,Chung2022ScalingIL}.
We explore zero-shot generalization to similar tasks in new domains.
Domain \textit{adaptation} in VQA has been explored, first by \cite{Xu2020OpenEndedVQ,Chao2018CrossDatasetAF} and most recently by \cite{Zhang2021DomainrobustVW}.
These fall into the general line of \textit{feature adaptation} methods for domain adaptation, as they align domain features.
Our method is more similar to pseudolabeling based methods for domain adaptation \cite{Kumar2020UnderstandingSF,Liu2021CycleSF} with the difference being that our pseudolabels are natural language rather than distributions.
Moreover, we do not focus on \textit{adaptation}, but zero-shot generalization.

\textbf{Visual Question Generation} is a well-explored topic with a long history of prior work \cite{Krishna2019InformationMV,Mostafazadeh2016GeneratingNQ,Zhang2017AutomaticGO,Li2018VisualQG}.
In contrast to prior work, our VQG teacher model \textit{does not} rely on or need paired ground truth annotations for an unlabeled image to generate questions.
SimpleAug \cite{Kil2021DiscoveringTU} and GuidedVQG\cite{Vedd2022GuidingVQ} relies on annotations such as bounding boxes to generate new questions, and requires pretrained object detectors, which have a limited label space.
WeaQ \cite{Banerjee2021WeaQAWS} requires captions to already be present, as does \cite{Changpinyo2022AllYM}, which additionally uses a large language model (T5-XXL with 11B parameters) to generate questions.
One similarity of our approach to \cite{Changpinyo2022AllYM} is that we both seek to use knowledge in a large model to generate questions, with the main differences being that we do not require ground-truth captions for unlabeled images, and we use a large vision-language model than a large language model.
VQAPG \cite{Yang2021DiversityAC} is similar to our approach in not requiring any ground-truth annotations, but focuses on creating a joint question-generation and question answering model that is consistent, rather than self-training a model with unlabeled data.
The authors of \cite{kafle-etal-2017-data} propose a VQG method that does not rely on ground-truth annotations, but their method is LSTM-based, rather than based on self-training with a large vision-language model.

\textbf{Self-Training} uses labeled data to train a teacher model.
The teacher model provides labels for auxiliary unlabeled data. 
Finally, a student model is trained on the labeled data augmented with newly-labeled data.
Previous work in self-training for computer vision focuses on image-classification \cite{Xie2020SelfTrainingWN,Yalniz2019BillionscaleSL} or object detection \cite{DBLP:conf/nips/ZophGLCLC020,9577846,4129456,Li2020ImprovingOD}.
A significant difference between classical self-training and our setting is that in the more traditional settings, the teacher and student have the same task.
In our setting, the task of the teacher (ask a question) is different than the task of the student (answer a question).
More similar to us, \cite{sachan-xing-2018-self} uses self-training for question-answering.
However, the teacher model of \cite{sachan-xing-2018-self} has a fundamentally different task, since it is a reading comprehension task, where the ground-truth answer is mentioned within the passage itself.
In our task, the teacher model must generate the ground-truth answer from its own internal knowledge and by inspecting an image.
\section{Method}
\begin{figure*}
    \centering
    \includegraphics{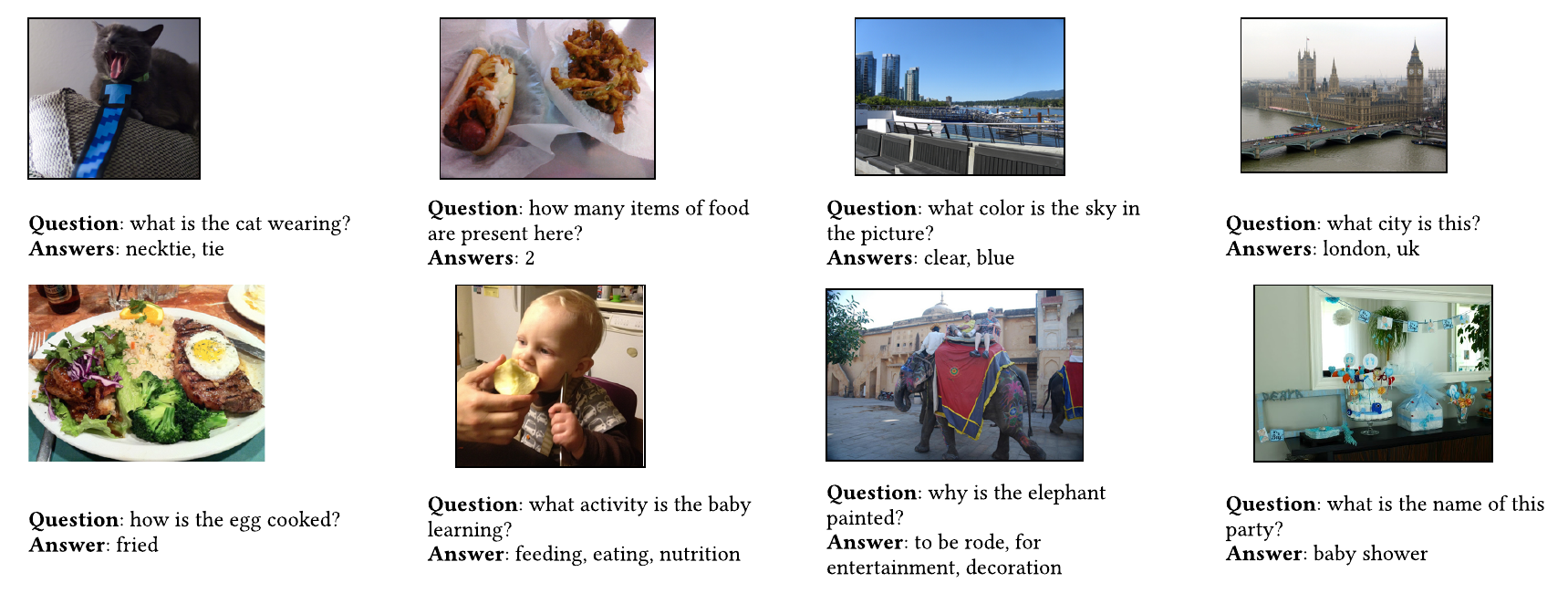}
    \caption{Example questions and answers generated by the teacher on unlabeled images. The questions include unusual pairings (cat wearing necktie) or require broad knowledge (identifying a baby shower or London landmarks) and inferences about scenes (the baby is learning).}
    \label{fig:qualitative-examples}
\end{figure*}
Our goal is to pseudolabel an unlabeled image $I$ with a generated question-answer pair $(Q,A)$ using a teacher (initialized from the VLM), and then train a student model (the initial VLM) on the real VQA pairs augmented with the generated VQA pairs.
To generate the pseudolabels, we first learn a visual question generation model on the real question-answer pairs and images as the teacher.
We denote this model VQG$_\mathrm{IC}$ to highlight the \textit{image-conditional} nature of the model, because the model generates both a question and answer conditional on an image alone. 
This approach is end-to-end, requires \textit{no ground truth annotations, bounding boxes, or handcrafted guidance}, and provides a generative model approximating $P(Q,A|I)$ that we can sample from.
We then feed the teacher model unlabeled images and stochastically decode from the teacher model to generate pseudolabels, which we parse into question answer pairs.
After the real samples in the dataset have been augmented with the self-generated samples, VQA training can proceed as normal.
Our approach is compatible with any modern encoder-decoder multimodal architecture. 
This is because our approach relies entirely on direct image-to-text generation, which is possible in modern large vision language models since their autoregressive decoders are designed to produce text conditioned on an image.
\subsection{The Teacher: Direct Image-Conditional VQG}
\label{sec:teacher}
Self-training requires a teacher model to produce pseudolabels that the student model then learns to mimic. 
In order to use unlabeled data for VQA, the teacher model must be able to pose a question and provide an answer given an unlabeled image, which is a different task from VQA.
Given an image $I$, a question $Q$ and answer $A$, the VQA student must approximate $P(A \mid Q,I)$, while the teacher model must approximate $P(Q,A \mid I)$.
Previous approaches to visual question generation (VQG) cannot work with unlabeled data because they approximate $P(Q\mid I,A)$, that is, they generate a question conditional on the image and a potential answer.
In contrast to these previous, answer-conditional VQG approaches, we develop an \textit{image-conditional} approach (VQG$_{IC}$) that we use as a teacher model.
Our approach also contrasts with self-training in image classification or object detection, which benefit from having the teacher and student \textit{both} approximating and predicting identically structured distributions $P(Y|I)$, where $Y$ is often a distribution over a (finite) label space.

To create the VQG$_{IC}$ teacher that approximates $P(Q,A | I)$, we treat the problem of learning such a model as a text-generation problem, and wish to train the autoregressive decoder of the vision-language model to approximate $P(T| I)$, where $T=(Q,A)$.
Let $\mathcal{D}_{QA}$ be a question-answer dataset we wish to create a teacher from.
For a sample $(Q, A, I) \in \mathcal{D}_{QA}$, we transform it into a target sequence of tokens $y_{1:N} = (y_1, y_2, \ldots y_n)$ by entering $(Q,A)$ into a structured template of the form  ``\textbf{Question}: \texttt{<question>}\textbf{?} \textbf{Answer}: \texttt{<answer>.}'' where \texttt{<question>} and \texttt{<answer>} are replaced by the content of $Q$ and $A$ respectively.
Once $y_{1:N} = (y_1, y_2, \ldots y_n)$ is obtained, we train the model by optimizing
\begin{equation}
\mathcal{L}_{\mathrm{VQG}}=-\sum_{n=1}^N \log P_\theta\left(y_n \mid y_{<n}, x\right)
\end{equation}
over all question-image-answer pairs in $\mathcal{D}_{QA}$, where $x$ is the latent encoded features in the standard encoder-decoder architecture and $\theta$ represents the VLM parameters.
The VQG$_\mathrm{IC}$ thus learns to maximize the conditional likelihood of a question-answer \textit{pair} represented as a unified string, given an image.
Recall that VQG$_{IC}$ is initialized from the parameters of an autoregressive VLM.
The VLM is a quality approximator of $P(T|I)$, having been exposed to a diverse number of images and paired text.
The VQG$_{IC}$ teacher can tap into this reservoir of knowledge, because a pseudo question-answer pair $(Q^\prime, A^\prime)$ is generated jointly as a text $T^\prime$, allowing us to sample from $P(T|I)$.

\subsection{Training the Student with Unlabeled Data}
\label{sec:pseudolabeling}
Once the VQG$_\mathrm{IC}$ teacher model has been obtained, self-training with unlabeled data can proceed.
To produce a pseudolabel $(Q^\prime, A^\prime)$ for an unlabeled image $I_{u}$, we first obtain $\mathbf{L}_{1:N}=\text{VQG}_{\mathrm{IC}}(I_u)$, where $\mathbf{L}_{1:N}$ are the logits of the decoder. The logits $\mathbf{L}_{1:N}$ define a distribution $P\left(L_N \mid L_{1: N-1}\right)$ over the tokens of the model's natural language vocabulary.
We then apply nucleus sampling \cite{Holtzman2020TheCC} to stochastically decode a text $T^\prime$ from $P\left(L_N \mid L_{1: N-1}\right)$.
The structured format of the generation template can then be easily parsed by a regular expression to recover a pseudo-question-answer pair $(Q^\prime, A^\prime)$ from the decoded text $T^\prime$.
This pair $(Q^\prime, A^\prime) = T^\prime$ is a sample from $P(T|I)$, and reflects textual knowledge about the content of an image known to the VLM.

We then proceed to pseudolabel the desired number of images and obtain any number of triplets of the form $(Q^\prime, A^\prime, I_{u})$, representing self-generated training data $\mathcal{D}^\prime_{QA}$in the style of a target dataset $\mathcal{D}_{QA}$.
We then augment the real dataset $\mathcal{D}_{QA}$ with the self-generated question-answer pairs on unlabeled images $\mathcal{D}^\prime_{QA}$ to create a self-augmented training dataset $\mathcal{D}_{\mathrm{Aug}QA} = \mathcal{D}^\prime_{QA} \cup \mathcal{D}_{QA}$.
The teacher model is no longer needed, and the student can be initialized from the checkpoint obtained after large-scale pretraining that the teacher model was initialized from.
At this point, VQA training can proceed as normal.
In our setting, we use the training procedure of BLIP \cite{blip} in which VQA is treated as an open-ended generation task, and the VQA objective can be expressed as the standard language modeling loss
\begin{equation}
\mathcal{L}_{\mathrm{VQA}}=-\sum_{n=1}^N \log P_\theta\left(y_n \mid y_{<n}, x_n\right)
\end{equation}
where $x_n$ is the $n$-th element of the multimodal sequence embeddings $\mathbf{X}_{1:N}$ produced by $\mathrm{VLM}(Q,I; \theta))$, $Q,I$ are the question and image, $y_{1:N}$ is the sequence of answer tokens, and $\theta$ represents the VLM parameters, which we initialize from the \textit{pretrained} weights rather than the teacher.
Why can high quality pseudolabels $(Q^\prime, A^\prime)$ be generated even when $\mathcal{D}_{QA}$ is small, and few pairs are available for adapting the teacher VQG$_{IC}$? 
Knowledge about the \textit{content} of the image in a textual form $P(T|I)$ is already well-learned by the VLM from which we initialize VQG$_{IC}$.
Thus, $\mathcal{D}_{QA}$ only needs sufficient pairs to teach VQG$_{IC}$ how to construct annotations matching the style of $\mathcal{D}_{QA}$.
\section{Experiments}

\noindent \textbf{Experimental Setup} We implement our framework in PyTorch\cite{NEURIPS2019_9015} and use the same hyperparameter settings for all experiments.
Our settings are taken from \cite{blip}.
We train each VQA model for 10 epochs, using the AdamW\cite{Loshchilov2017FixingWD} optimizer with a weight decay of 0.05 and a linear LR decay to 0 from an initial LR 2e-5.
Each VQG model is trained for 10 epochs with the same weight decay and an initial LR of 2e-5.
For VQA, we use a global batch size of 64 on 4 GPUs, with a per device batch size of 16. 
For VQG, we use a global batch size of 128, with a per device batch size of 32.
All models are initialized from pretrained BLIP\cite{blip} checkpoints.
For VQA, we use an image size of $480\times480$ and an image size of $384\times384$ for VQG.
For all datasets, we use the official training, validation, and test splits.

\noindent\textbf{Baseline} As a strong baseline model, we use the ViT-B/16 version of the BLIP \cite{blip} model pretrained on 129M image-text pairs.
BLIP\cite{blip} has an autoregressive decoder and is trained for text-generation, making it easy to adapt to text-generation tasks.
When decoding, we use nucleus sampling with a top-$p$ of 0.92.
Additional experiments and visualizations can be found in the supplemental material.
\subsection{Self-Training: A-OKVQA \& ArtVQA}
\begin{table}[]
\centering
\begin{tabular}{@{}llll@{}}
\toprule
 &  & \multicolumn{2}{c}{A-OKVQA} \\ \cmidrule(l{0.5em}r{0.5em}){3-4}
    & Model   & Validation & Test \\ \midrule
(a) & ViLBERT \cite{vilbert} & 49.1       & 41.5\\
(b) & LXMERT \cite{lxmert}  & 51.4       & 41.6 \\
(c) & KRISP \cite{krisp}   & 51.9       & 42.2   \\
(d) & GPV-2 \cite{gpv2}  & 60.3       & 53.7 \\
(e) & BLIP \cite{blip}   & 57.1       &        \\
(f) & BLIP$_{\mathrm{VQAv2}}$ \cite{blip}   & 67.8       & \textbf{59.5}        \\
\midrule
(g) & BLIP + \stda  & 62.1       & 54.5                   \\
\multicolumn{2}{l}{\% gain w.r.t baseline}      & \textcolor{Green}{+5.0}        &  \\ 
\multicolumn{2}{l}{\% gain w.r.t best prior work}      & \textcolor{Green}{+1.8}        & \textcolor{Green}{+0.8} \\ 
(h) & BLIP$_{\mathrm{VQAv2}}$ + \stda  & \textbf{68.9}       & \textbf{59.5}                   \\
\multicolumn{2}{l}{\% gain w.r.t baseline}      & \textcolor{Green}{+1.1}        & \textcolor{Green}{+0.0} \\ 
\multicolumn{2}{l}{\% gain w.r.t best prior work}      & \textcolor{Green}{+8.6}        & \textcolor{Green}{+5.8} \\ 
\bottomrule
\end{tabular}
\caption{
\stda~improves performance on knowledge-based VQA, even on a strong baseline pretrained on 129M pairs. 
}
\label{tab:aokvqa}
\end{table}
\begin{table}[]
\centering
\begin{tabular}{@{}llll@{}}
\toprule
&                             & \multicolumn{2}{c}{ArtVQA Accuracy} \\ \cmidrule(l{0.5em}r{0.5em}){3-4}
& Model                        & Overall           & Grounded           \\ \midrule
(a) & BAN \cite{Kim2018BilinearAN}                         & 22.4              & -                           \\
(b) & BLIP \cite{blip}                        & 21.36             & 81.71                       \\
(c) & VIKING \cite{aqua}                      & 55.5              & 78.74                       \\
(d) & VIKING$_\mathrm{VLM}$                   & 55.9              & 81.9                        \\ \midrule
(e) & BLIP + \stda                     & 21.68             & \textbf{83.86}                       \\
& \% gain w.r.t baseline & \textcolor{Green}{+0.32}               & \textcolor{Green}{+2.15}                        \\
(f) & VIKING$_\mathrm{VLM}$ + \stda             & \textbf{56.86}             & \textbf{83.86}                       \\
& \% gain w.r.t baseline & \textcolor{Green}{+0.92}              & \textcolor{Green}{+1.96}                        \\ \bottomrule
\end{tabular}
\caption{\stda~improves VQA on fine art images\cite{aqua} for VIKING and BLIP models. Grounded denotes visually grounded questions.}
\label{tab:artvqa}
\end{table}
\begin{table*}[]
\centering
\begin{tabular}{@{}lccc|c@{}}
\toprule
Question Type         & Well-Posed Question & Answers Correct & Answerable    & \% of Total (95\% CI) \\ \midrule
External Knowledge    & 73\%                  & 62\%              & 70\%            & 29.6\% - 50.00\%            \\
Visual Identification & 94\%                  & 88\%              & 94\%            & 11.18\% - 27.65 \%         \\
Visual Reasoning      & 83\%                  & 70\%              & 80\%            & 32.54\% - 53.17\%         \\
\midrule
Overall (95\% CI)     & 71.16\% - 87.96\%       & 59.77\% - 78.98\%   & 68.83\% - 86.22\% &                       \\ \bottomrule
\end{tabular}
\caption{We manually inspect 100 questions and answers generated by the teacher model finetuned on A-OKVQA. 
We show the 95$\%$ confidence interval obtained by a proportion test. 
Annotator agreement on A-OKVQA is about 79.5\% on the validation set.
}
\label{tab:synth-question-stats}
\end{table*}
\begin{figure*}
    \centering
    \includegraphics{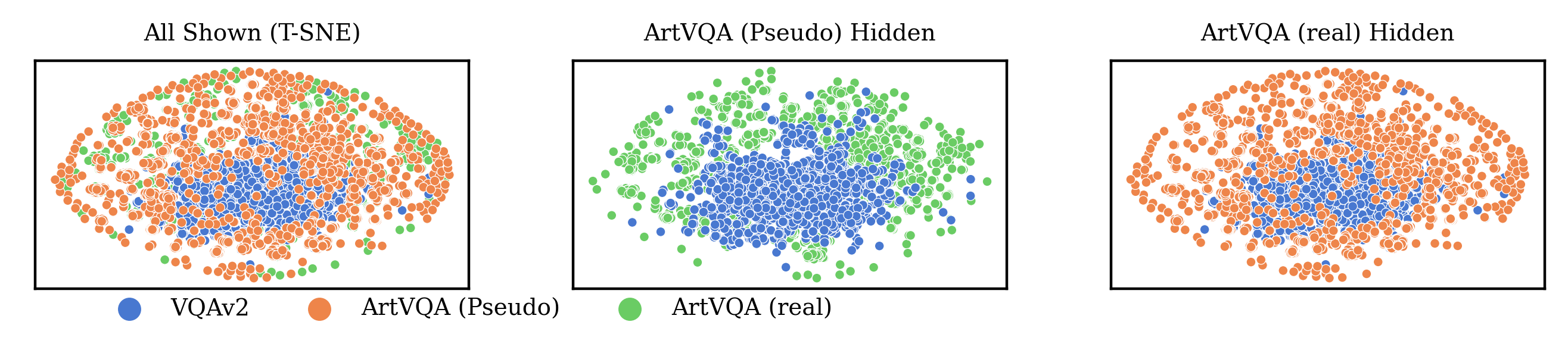}
    \caption{A T-SNE embedding shows that questions generated by a teacher finetuned on ArtVQA (\textcolor{orange}{orange}) differ from real VQAv2 questions (\textcolor{RoyalBlue}{blue}) and are more similar to the real ArtVQA questions (\textcolor{Green}{green}), yet more diverse, covering a larger area. We use SimCSE\cite{Gao2021SimCSESC} to obtain a dense vector representation of each sentence. All the sets of questions are embedded together with T-SNE.} 
    \label{fig:tsne}
\end{figure*}
\begin{figure}
    \centering
    \includegraphics[width=\linewidth]{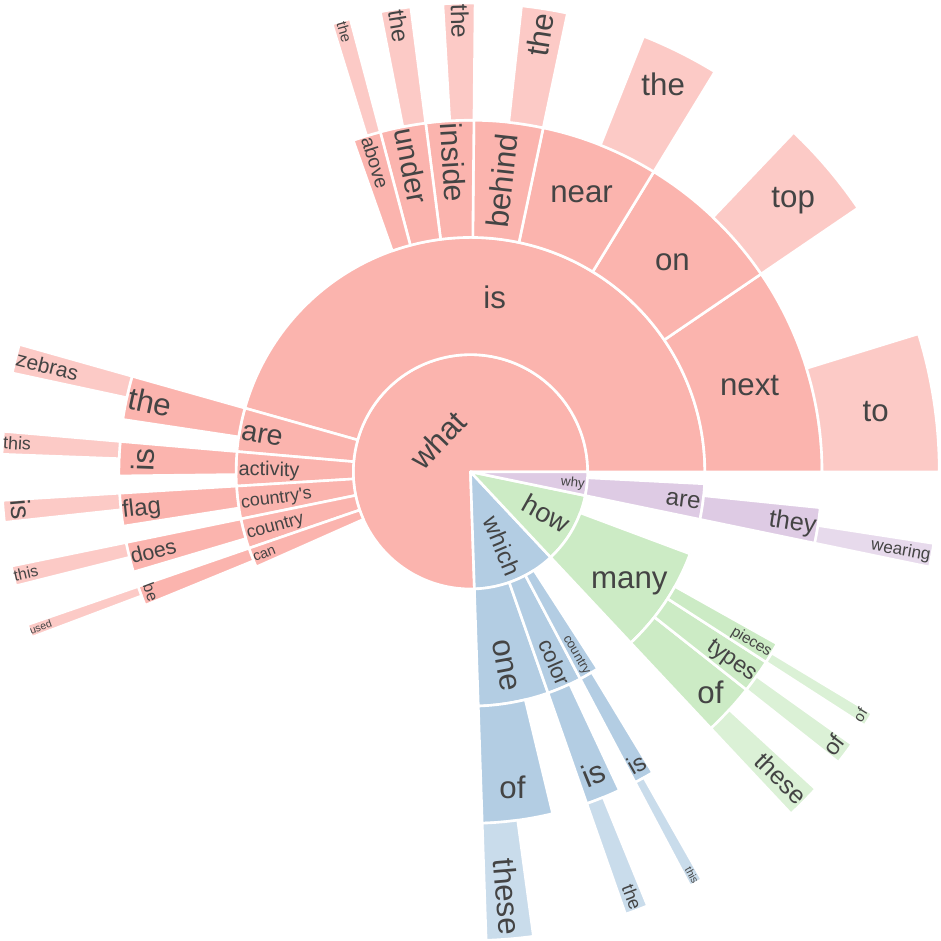}
    \caption{Sunburst chart of questions generated by a teacher model finetuned on A-OKVQA.}
    \label{fig:sunburst}
\end{figure}
We evaluate \stda~in two domains: outside knowledge VQA on natural images with A-OKVQA \cite{AOKVQA} and outside knowledge VQA on fine-art images with AQUA \cite{aqua}.
We use the COCO 2017 unlabeled set \cite{Lin2014MicrosoftCC} as a source of additional images for A-OKVQA, and SemArt \cite{garcia2018how} as a source of fine art images for ArtVQA.
On A-OKVQA, we perform model selection over students trained with varying amounts of \stda~with the training set, and on ArtVQA, we use the validation set.
On A-OKVQA (Table \ref{tab:aokvqa}), we show that self-taught data augmentation improves overall performance, especially in the setting where no extra data (VQAv2) is available.
BLIP with \stda~achieves SOTA performance on A-OKVQA without transfer learning (row g in Table \ref{tab:aokvqa}), even relative to competitors using transfer learning.
This performance improvement holds even when $447k$ \textit{real} pairs from VQAv2 are used for transfer learning, suggesting that self-taught data augmentation offers real improvements over manual annotations.
On fine art VQA (Table \ref{tab:artvqa}), we show that self-taught data augmentation achieves state-of-the-art and improves overall performance, with a large increase for visually grounded questions.
\subsection{Ablations \& Analysis of Pseudolabels}
We manually evaluate 
100 randomly sampled questions generated by the teacher model on A-OKVQA (Table \ref{tab:synth-question-stats}).
The generated questions and answers are noisier than the real questions and answers, but the levels of noise are not substantially below the human agreement on A-OKVQA.
Questions which require visual reasoning or external knowledge are harder to generate correctly compared to those that require simpler visual identification (e.g. ``what is this object?'').
Next, we show using t-SNE\cite{JMLR:v9:vandermaaten08a} that the teacher model learns to copy the ``style'' of questions in a particular dataset (Fig \ref{fig:tsne}).
Synthetic questions generated by a teacher finetuned for a specific dataset (ArtVQA) are more similar to the style of the questions found in the target dataset compared to real questions from a different dataset (VQAv2), while being more diverse.
\begin{table*}[]
\centering
\adjustbox{max width=\linewidth}{
\begin{tabular}{@{}llllllllc@{}}
\toprule
\multicolumn{2}{c}{Images} & \multicolumn{3}{c}{Questions} &  &  &  &  \\ \cmidrule(l{0.5em}r{0.5em}){1-2} \cmidrule(l{0.5em}r{0.5em}){3-5}
Labeled  &  Unlabeled & Real & Synthetic & Total & Multiplier & Accuracy & \% Gain & Questions/Image\\ \midrule
17,000 & 0               & 17,000  & 0         & 17,000   & 1x (baseline)        & 57.11 &     & N/A \\
17,000 &0               & 17,000  & 17,000      & 34,000   & 2x         & 57.85 & \textcolor{Green}{+0.74}    &  1 / 1\\
17,000 &0               & 17,000  & 34,000       & 51,000   & 3x         & \textbf{60.01} & \textbf{\textcolor{Green}{+2.90}}  & 2 / 1\\
17,000 &0               & 17,000  & 51,000       & 68,000   & 4x         & 59.73 &  \textcolor{Green}{+2.62}   & 3 / 1\\ \midrule
17,000 &0              & 17,000  & 0         & 17k   & 1x (baseline)        & 57.11  &  &  N/A \\
17,000 &8,500              & 17,000  & 17,000       & 34,000   & 2x         & 60.69   & \textcolor{Green}{+3.57}  & 2 / 1 \\
17,000 &17,000              & 17,000  & 34,000       & 51,000   & 3x         & \textbf{62.09}  &  \textbf{\textcolor{Green}{+4.98}}  & 2 / 1\\
17,000 &25,500              & 17,000  & 51,000       & 68,000   & 4x         & 61.31    &  \textcolor{Green}{+4.20}    & 2 / 1\\ \bottomrule
\end{tabular}
}
\caption{
\stda~can improve performance even without additional unlabeled images, by generating more QA pairs for already labeled images.
However, using previously unlabeled and unseen images results in further improvements. A-OKVQA is used.
}
\label{tab:ablation}
\end{table*}
\begin{table*}
   \begin{tabular}{@{}llllllcll@{}}
\toprule
    &                \multicolumn{3}{c}{\# of Real + Synthetic QA Pairs}                                &  \multicolumn{3}{c}{Robustness Test Sets} &  &  \\ \cmidrule(l{0.5em}r{0.5em}){5-7}  \cmidrule(l{0.5em}r{0.5em}){2-4}
    & Real     & Synthetic     & Multiplier    & AdVQA & VQA-CE & VQA-Rephrasings & Avg. \% Increase & Robustness Total \\ \midrule
(a) & 17,000          & 0                  & $\times$1             & 31.06 & 51.43  & 65.88                      & 0                & 148.37                 \\
(b) & 17,000          & 2,000               & $\times$1.1           & 37.09 & 52.96  & 67.94                    & \textcolor{Green}{+3.21}             & 157.99                 \\
(c) & 17,000          & 4,500               & $\times$1.3           & 36.99 & 53.15  & \textbf{67.98}           & \textcolor{Green}{+3.25}             & 158.12                 \\
(d) & 17,000          & 8,000               & $\times$1.5           & 37.34 & \textbf{53.33}  & 67.57           & \textcolor{Green}{+\textbf{3.29}}             & \textbf{158.24 }                \\
(e) & 17,000          & 12,000              & $\times$1.7           & \textbf{37.43} & 52.62  & 67.35           & \textcolor{Green}{+3.01}             & 157.4                  \\
(f) & 17,000          & 17,000              &$\times$2             & 36.95 & 52.05  & 66.95                    & \textcolor{Green}{+2.53}             & 155.95                 \\
(g) & 17,000          & 34,000              & $\times$3             & 36.89 & 51.00     &      65.64                 &      \textcolor{Green}{+1.72}            & 153.53                  \\
(h) & 17,000          & 51,000              & $\times$4             & 36.06 & 50.25  &      64.78                    &       \textcolor{Green}{+0.91}           & 151.09                 \\
\midrule
    & \multicolumn{3}{l}{Max \% increase on each dataset} & \textcolor{Green}{+6.03}  & \textcolor{Green}{+1.9}    & \textcolor{Green}{+2.1}             &                  &             \textcolor{Green}{+9.87}           \\ \bottomrule
\end{tabular} 
\caption{\stda~improves robustness of VQA models on AdVQA (adversarially searched questions), VQA-CE (multimodal shortcut learning) and VQA-Rephrasings test sets. The baseline (a) is trained on VQAv2 after pretraining, then finetuned on A-OKVQA.}
\label{tab:robustness}
\end{table*}
We show that the performance gains of \stda~are due to novel-question answer pairs (first half of Tab \ref{tab:ablation}) that add information not present in the ground-truth QA pairs, not only due to the additional images.
However, the student model benefits from \textit{both} the novel-question answer pairs and unlabeled images (second half of Table \ref{tab:ablation}).

\textbf{Optimal Amount of Augmentation} We explore how the amount of augmentation affects performance.
The highest performance on the A-OKVQA validation and test sets is reached when the number of synthetic is double that of the real pairs (Table \ref{tab:ablation}).
When transfer learning from VQAv2, the ratio is different, and peak performance is reached when the number of synthetic pairs is $50\%$ the number of real pairs (Table \ref{tab:robustness},\ref{tab:ablation}).
Performance and robustness improvements (Table \ref{tab:robustness}) saturate as increasing amounts of synthetic pairs are added, which may be the result of task-irrelevant information seeping into the dataset due to stochastic sampling.
\subsection{Robustness}
We investigate whether the self-taught data augmentation improves robustness of VQA models.
We consider three known weaknesses.
The first is adversarially searched questions, collected in the AdVQA \cite{advqa} dataset through human-in-the-loop attacks against state-of-the-art VQA models.
In Table \ref{tab:robustness}, we show that models trained with self-taught data augmentation perform significantly better ($20\%$ relative improvement and $6\%$ absolute improvement) on AdVQA.
The second form of robustness we consider is resistance to multimodal shortcut learning, which the VQA-CE (Counterexamples) \cite{vqa_ce} test set measures.
The test set is constructed so that models which have learned to answer questions using shortcuts based on correlations in the VQAv2 training set (ex: tennis racket detected  + question about sport $\rightarrow$ always answer tennis) will display reduced performance on the VQA-CE test set. 
We construct our A-OKVQA models by transfer learning from the VQAv2 training set, so VQA-CE can be used to test multimodal shortcut learning in our models.
In Table \ref{tab:robustness}, we show that models trained with self-taught data augmentation are more resistant to shortcut learning ($1.9\%$ absolute improvement on VQA-CE) compared to the baseline model trained without self-taught data augmentation. 
Finally, we consider robustness to rephrasings.
VQA models have been shown to be inconsistent when evaluated on rephrasings \cite{vqa_rephrasings}.
The VQA-Rephrasings test set consists of 3 human-provided rephrasings of the questions in the VQAv2 test set, intended to test the robustness of the model to rephrasings.
On VQA-Rephrasings, self-taught data augmentation induces a $2.1\%$ performance improvement relative to the baseline model, though both the baseline model and augmented models were initialized from from the same weights learned on the VQAv2 training set prior to finetuning on A-OKVQA. 
\subsection{Domain Generalization}
We hypothesize that self-taught data augmentation may improve domain generalization, because the student model has been exposed to a greater diversity of questions and answers.
To test this, we compare the generalization of the baseline model and models trained with self-taught data augmentation on unseen test sets from three different domains.
Concretely, we treat the natural-image based A-OKVQA task as the source task, and evaluate on VQA datasets from three target domains: medical, fine art, and remote sensing.
For medical VQA, we use the PathVQA \cite{he2020pathvqa} dataset containing question and answers on pathology images.
For fine art, we used the previously described AQUA \cite{aqua} dataset for visual question answering on art images.
For remote sensing, we use the RSVQA dataset \cite{Lobry2020RSVQAVQ}, containing question and answers on satellite images.
We display the results in Table \ref{tab:domain-generalization}.
Across all three domains, self-taught data augmentation improves domain generalization over the baseline model.
The improvement is greatest on fine art images, as the fine art domain is closest to the natural image domain with respect to the images, questions, and answers.
\subsection{Numerical Reasoning}
Numerical reasoning is required to answer questions such as ``how many sheep are looking at the camera''. 
Naive transfer learning from VQAv2 to A-OKVQA results in catastrophic forgetting of numerical reasoning, and naive finetuning on A-OKVQA results in models with poor numerical reasoning.
In Table \ref{tab:0-shot-tranfer}, we show that \stda~significantly aids numerical reasoning when finetuning on a small-scale VQA dataset such as A-OKVQA.
We measure numerical reasoning using questions labeled as requiring numerical answers on VQAv2 and the VQA-Rephrasings datasets.
When transfer learning from VQAv2 (first half of Table \ref{tab:0-shot-tranfer}), self-taught data augmentation results in an absolute increase of $29.81$\% and $24.71$\% on numerical questions on VQAv2 and VQA-Rephrasings.
When finetuning directly on A-OKVQA (2nd half of Table \ref{tab:0-shot-tranfer}), self-taught data augmentation results in an absolute increase of $3.63$\% and $10.57$\%.
These results suggest that self-taught data augmentation can prevent catastrophic forgetting of numerical reasoning when transfer learning, and improve numerical reasoning significantly, even when the dataset used to train the teacher model has few numerical reasoning questions.
One reason for this is that the the word ``how'' is a high-probability word to start a question with, and is naturally followed by ``many'' (Fig \ref{fig:sunburst}) resulting in numerical questions being generated. 
\begin{table}[]
\centering
\begin{tabular}{llll}
\hline
        & \multicolumn{3}{c}{Target (0-shot)}          \\ \cmidrule(l{0.5em}r{0.5em}){2-4}
Model  & ArtVQA & PathVQA & RSVQA \\ \midrule
Baseline (BLIP)  & 31.65 & 25.09  & 37.78\\
BLIP + \stda   & 38.03 & 26.76 & 38.99\\ \midrule
\% gain w.r.t baseline & \textcolor{Green}{+6.38}   & \textcolor{Green}{+1.67}   & \textcolor{Green}{+1.1} \\ \bottomrule
\end{tabular}
\caption{\stda improves domain generalization from natural images (A-OKVQA) to art QA, medical QA, and remote sensing QA.}
\label{tab:domain-generalization}
\end{table}
\begin{table}
\adjustbox{max width=\linewidth}{
\begin{tabular}{@{}lllll@{}}
\toprule
          &      \multicolumn{2}{c}{\# Training Pairs}             & \multicolumn{2}{c}{Numerical Reasoning}                \\  \cmidrule(l{0.5em}r{0.5em}){4-5} \cmidrule(l{0.5em}r{0.5em}){2-3}
Initialization      & Real  & Synth & VQAv2               & VQA-Rephrasings \\ \midrule
BLIP$_{VQAv2}$ & 17000 & 0         & 13.49               & 13.06           \\
BLIP$_{VQAv2}$  & 17000 & 2000      &  38.73                   & 33.74           \\
BLIP$_{VQAv2}$  & 17000 & 4500      & 40.4                & 35.91           \\
BLIP$_{VQAv2}$  & 17000 & 8000      & 42.9                & 36.5            \\
BLIP$_{VQAv2}$  & 17000 & 12000     & \textbf{43.3}                & \textbf{37.77}           \\ \midrule
\multicolumn{3}{c}{max \% gain w.r.t baseline}      &          \textcolor{Green}{+29.81}       &     \textcolor{Green}{+24.71}       \\ \midrule
BLIP      & 17000 & 0         & 1.42                & 1.29            \\
BLIP      & 17000 & 17000     & 4.53                & 11.44           \\
BLIP      & 17000 & 34000     & \textbf{5.05}                & 11.77           \\
BLIP      & 17000 & 51000     & 4.26                & \textbf{11.86}           \\ \midrule
\multicolumn{3}{c}{max \% gain w.r.t baseline}      &          \textcolor{Green}{+3.63}       &     \textcolor{Green}{+10.57}       \\ 
\bottomrule
\end{tabular}
}
\caption{
\stda~improves numerical reasoning when finetuning on a small-scale dataset (A-OKVQA).
BLIP$_{VQAv2}$ indicates transfer learning from VQAv2, and BLIP indicates direct finetuning.
}
\label{tab:0-shot-tranfer}
\end{table}

\section{Conclusion \& Future Work}
We present \stda, a framework for self-improving large VLMs on small-scale visual question answering tasks with unlabeled data.
The limitations of \stda~suggest several opportunities for further work.
First, the pseudo-QA pairs can be noisy. 
Combining \stda with methods for fact-checking based on external knowledge \cite{DBLP:journals/corr/abs-2112-09924}, logically consistent self-reasoning \cite{Jung2022MaieuticPL}, or chain-of-thought prompting \cite{Wei2022ChainOT} to rationalize answers may result in higher quality pairs for self-training.
Second, learning the teacher model may fail for specialized domains (e.g. medical), because the vocabulary is too specialized.
Third, biases in the VLM or pretraining data may be amplified by self-training, and addressing these biases may reduce multimodal shortcut learning.
Finally, self-training is yet to be explored with recently developed billion-parameter VLMs \cite{blip_2,magma}.

{\small
\bibliographystyle{ieee_fullname}
\bibliography{paper}
}

\clearpage
\appendix
\noindent In the supplementary material, we discuss the following:
\begin{enumerate}[label=(\Alph*)]
    \item T-SNE embeddings for PathVQA \cite{he2020pathvqa}
    \item Typology of Generated Questions
    \item Examples answered correctly / incorrectly on AdVQA \cite{advqa}
    \item Examples answered correctly/ incorrectly on ArtVQA \cite{aqua} and PathVQA \cite{he2020pathvqa}
\end{enumerate}
\section{T-SNE plots}
\begin{figure*}
    \centering
    \includegraphics{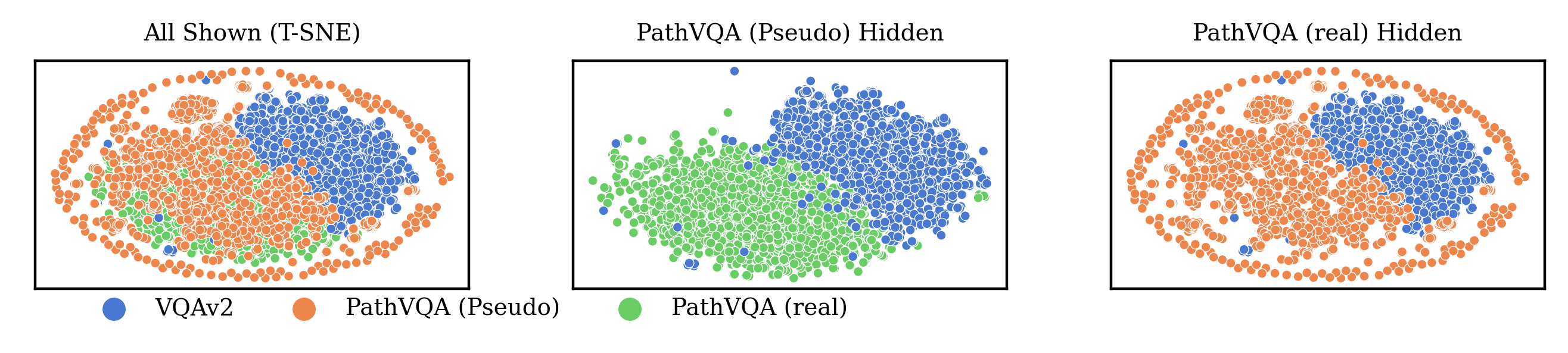}
    \caption{T-SNE plots of real and pseudo Path-VQA questions compared to VQAv2 questions. All questions were embedded together with SimCSE.}
    \label{fig:pathvqa-tsne}
\end{figure*}

In Fig \ref{fig:pathvqa-tsne}, we show a T-SNE embedding of real questions from VQAv2, real questions from PathVQA, and synthetic PathVQA-style questions generated by \stda.
Sentence-level dense representations of each questions were obtained with SimCSE, and were embedded together with T-SNE\footnote{We use \url{https://scikit-learn.org/stable/modules/generated/sklearn.manifold.TSNE.html}} using a cosine metric and a perplexity of 5. 
The same settings were used for the T-SNE table in the main plot. 
We randomly sample $1000$ questions from each dataset.
In Fig \ref{fig:pathvqa-tsne}, the separation between the real PathVQA questions and the VQAv2 questions can be clearly seen. 
The distribution of the synthetic PathVQA questions is similar to that of the real PathVQA questions, while being more diverse.
The synthetic PathVQA questions are also clearly differentiated from the VQAv2 questions.
\section{Typology of Generated Questions}
We manually inspect and categorize generated questions and answers from the A-OKVQA dataset according to the typology introduced in \cite{AOKVQA}.
We briefly describe the typology.
\begin{itemize}
    \item \textbf{Physical Reasoning} questions require knowledge about the physical world and processes that occur in it. An example of such knowledge is that shaded areas are cooler than unshaded areas in direct sunlight.
    \item \textbf{Visual Reasoning} questions require knowledge about the spatial relationships in an image. An example of such knowledge is knowing what is to the left or right of a specified object.
    \item \textbf{Numeracy} involves reasoning about numbers. The simplest incarnation of numeracy is counting, such as questions of the sort "how many sheep are present?". This category is not in the original typology introduced by A-OKVQA.
    \item \textbf{Knowledge Base} questions are those that deal with factoid, empirical knowledge.
    This includes knowledge such as knowing the names of rare categories, or the date that an event occurred. 
    \item \textbf{Commonsense} questions are those which involve understanding common human relationships and behavior. This is generally collective knowledge known to humans through experience, and may be culturally dependent. For example, suits are often worn to formal events, and certain foods are associated with specific meals.
\end{itemize}
We note that this typology is not exhaustive and the distinctions between some categories can be blurry. 
In the main paper, we use the category names ``Visual Identification'', ``External Knowledge'' and ``Visual Reasoning''. 
While there is no analogue for the ``Visual Identification'' category in the typology of A-OKVQA (because ``Visual Identification'' questions rarely require external knowledge), the category of ``External Knowledge'' subsumes Commonsense and Knowledge Base questions, while ``Visual Reasoning'' subsumes Visual Reasoning, Physical Reasoning, and Numeracy questions.

\begin{figure*}
    \centering
    \includegraphics{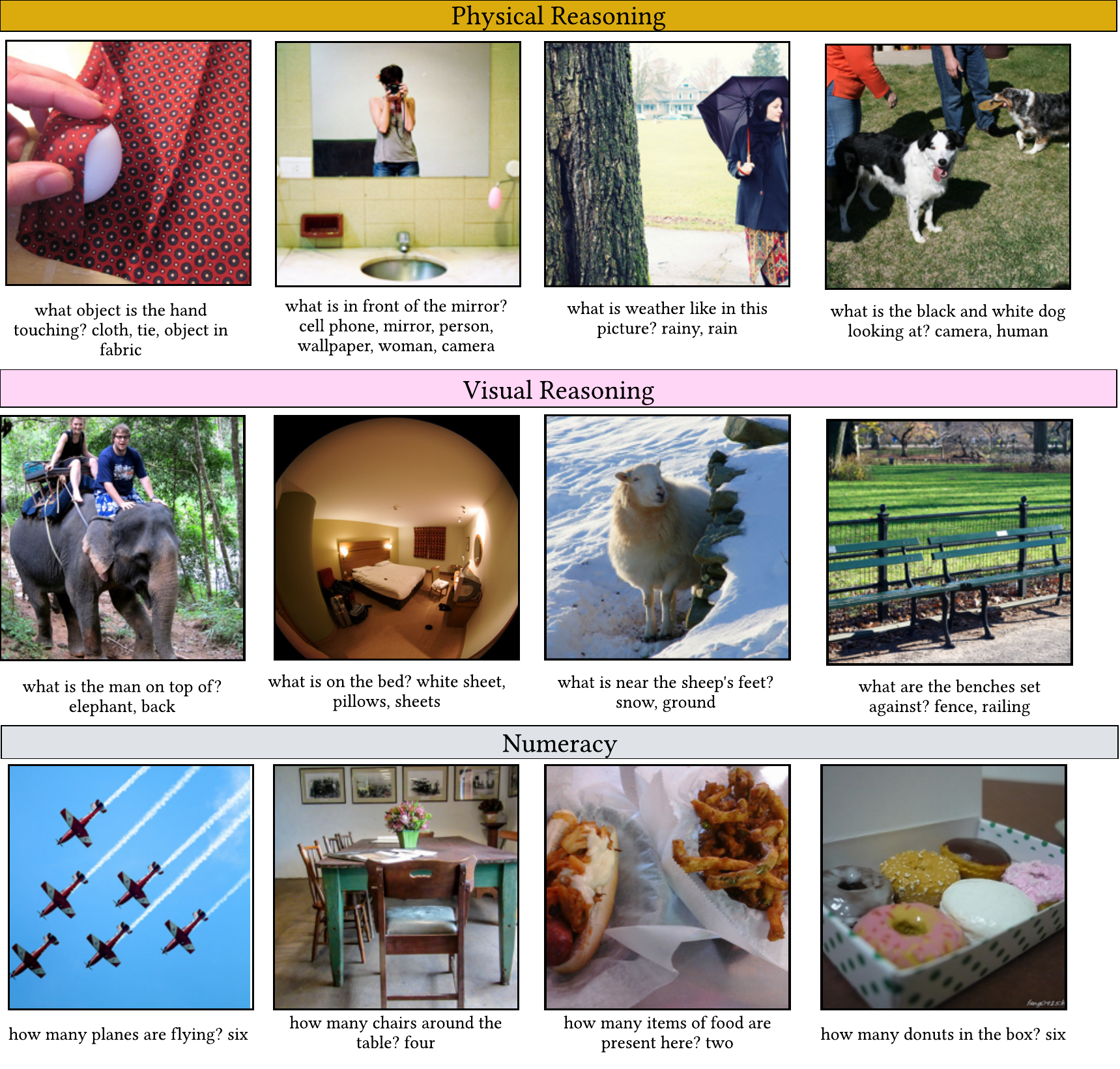}
    \caption{Examples of questions generated by \stda~involving physical reasoning, visual reasoning, or numeracy.}
    \label{fig:question-types-phys-vis-numeracy}
\end{figure*}

\begin{figure*}
    \centering
    \includegraphics{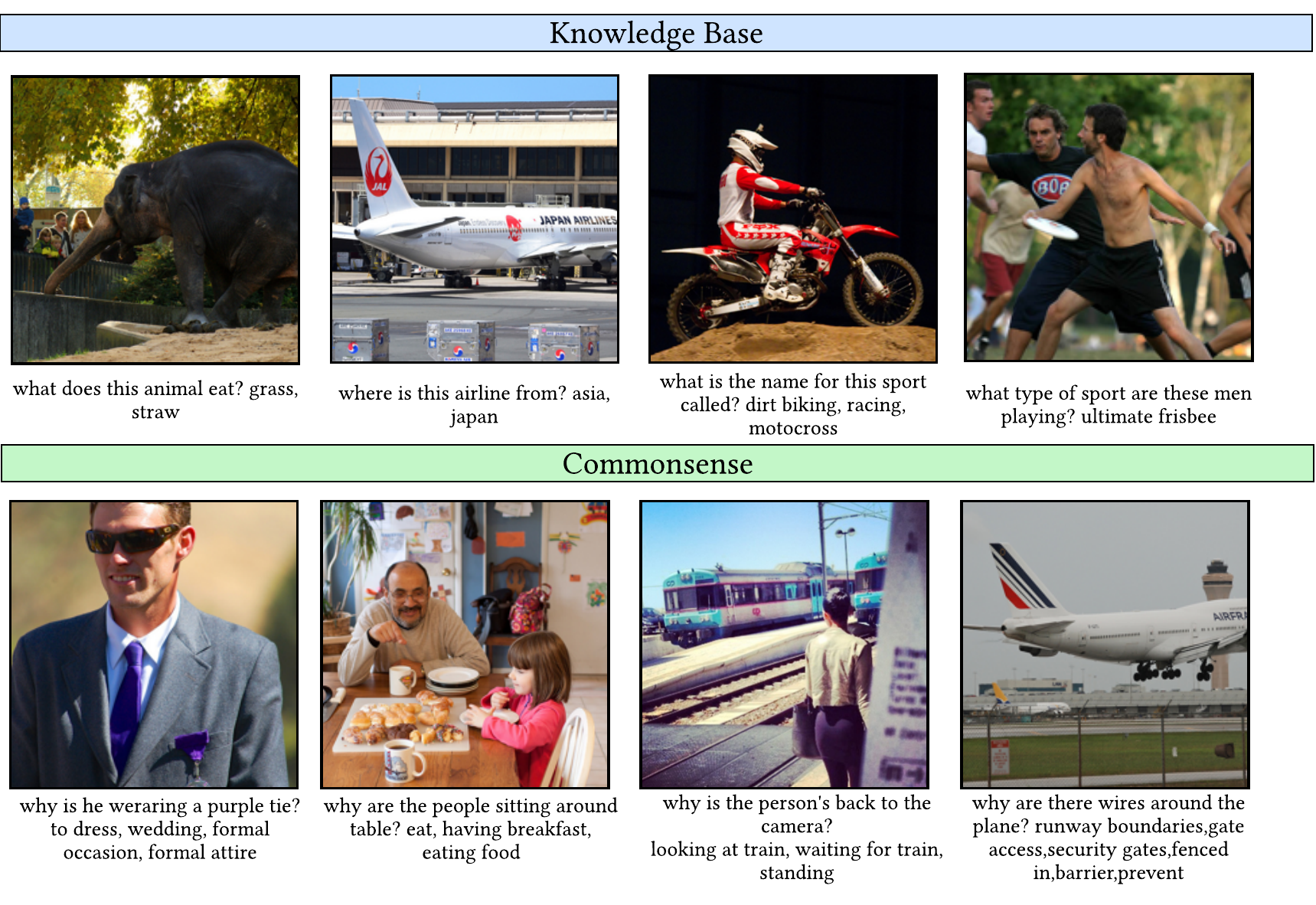}
    \caption{Examples of questions generated by \stda~involving knowledge base-type/external knowledge (e.g. factoids) and commonsense reasoning.}
    \label{fig:question-types-kb-commonsense}
\end{figure*}

In Fig \ref{fig:question-types-phys-vis-numeracy}, we show examples of Physical Reasoning, Visual Reasoning, and Numerical questions generated by the teacher model finetuned on A-OKVQA.
In Fig \ref{fig:question-types-kb-commonsense}, we show examples of Knowledge Base and Commonsense questions generated by the teacher model finetuned on A-OKVQA.
Existing datasets for general-purpose VQA tend to test Physical Reasoning, Visual Reasoning, and Numeracy. 
However, as shown in Fig \ref{fig:question-types-kb-commonsense}, large-scale pretraining endows a VLM with rich commonsense knowledge.
Surprisingly, the teacher can generate knowledge-base style questions, such as ``what does this animal eat?''. 
However, a VLM is limited by its size, and may not be able to store a large amount of factoid knowledge within its parameters.
Thus, incorporating external knowledge bases may be a useful next step for generating deeper questions.

\section{AdVQA Correct / Incorrect Samples}
\begin{figure*}
    \centering
    \includegraphics{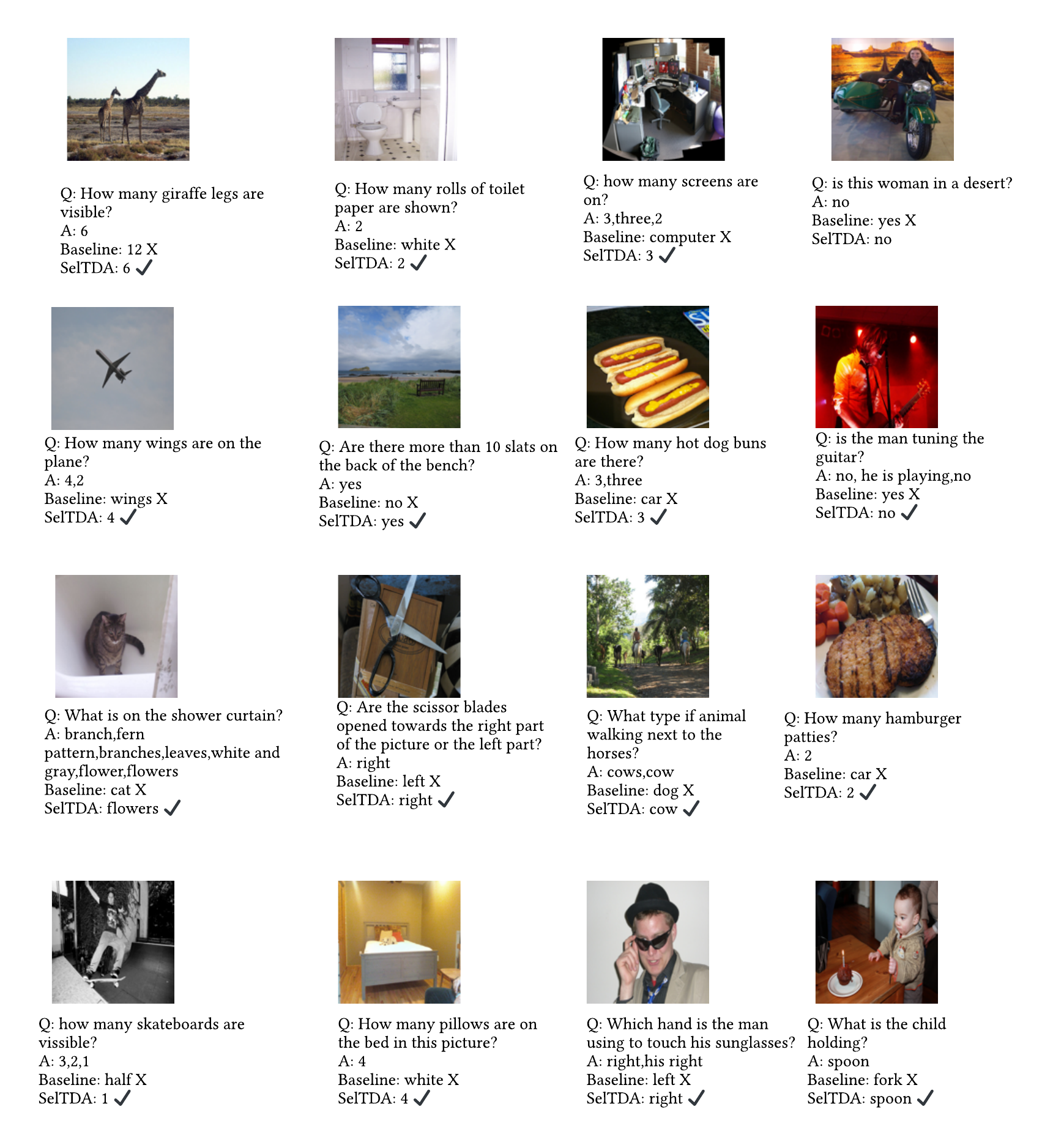}
    \caption{Examples of adversarial questions (AdVQA) that the baseline model gets wrong, but the model trained with \stda~gets correct.}
    \label{fig:advqa-seltda-right}
\end{figure*}

We next discuss results on the AdVQA dataset.
We show with qualitative examples how a model trained with \stda~improves over a baseline model, and what room for improvement remains.
In Fig \ref{fig:advqa-seltda-right}, we show examples of questions from the AdVQA dataset that a model trained with \stda~answers correctly, but a baseline model does not. 
We observe several weaknesses that \stda~helps mitigate.
First, note that \stda~improves counting. 
For some questions (Fig \ref{fig:advqa-seltda-right}, col 1, row 1) the model trained with \stda~answers the correct number, while the baseline model answers with the wrong number.
Another interesting pattern we observe is that \stda~trained models appear to parse and understand the questions better.
Consider row 2, col 1 and row 1, col 3 in Fig \ref{fig:advqa-seltda-right}. 
Both are straightforward questions with numerical answers: ``how many screens are on?'' and ``how many wings are on the plane?''
Surprisingly, the baseline model fails to parse the question correctly and commits a category error, producing a non-numerical answer (``computer'' and ``wing'' respectively).

\begin{figure*}
    \centering
    \includegraphics{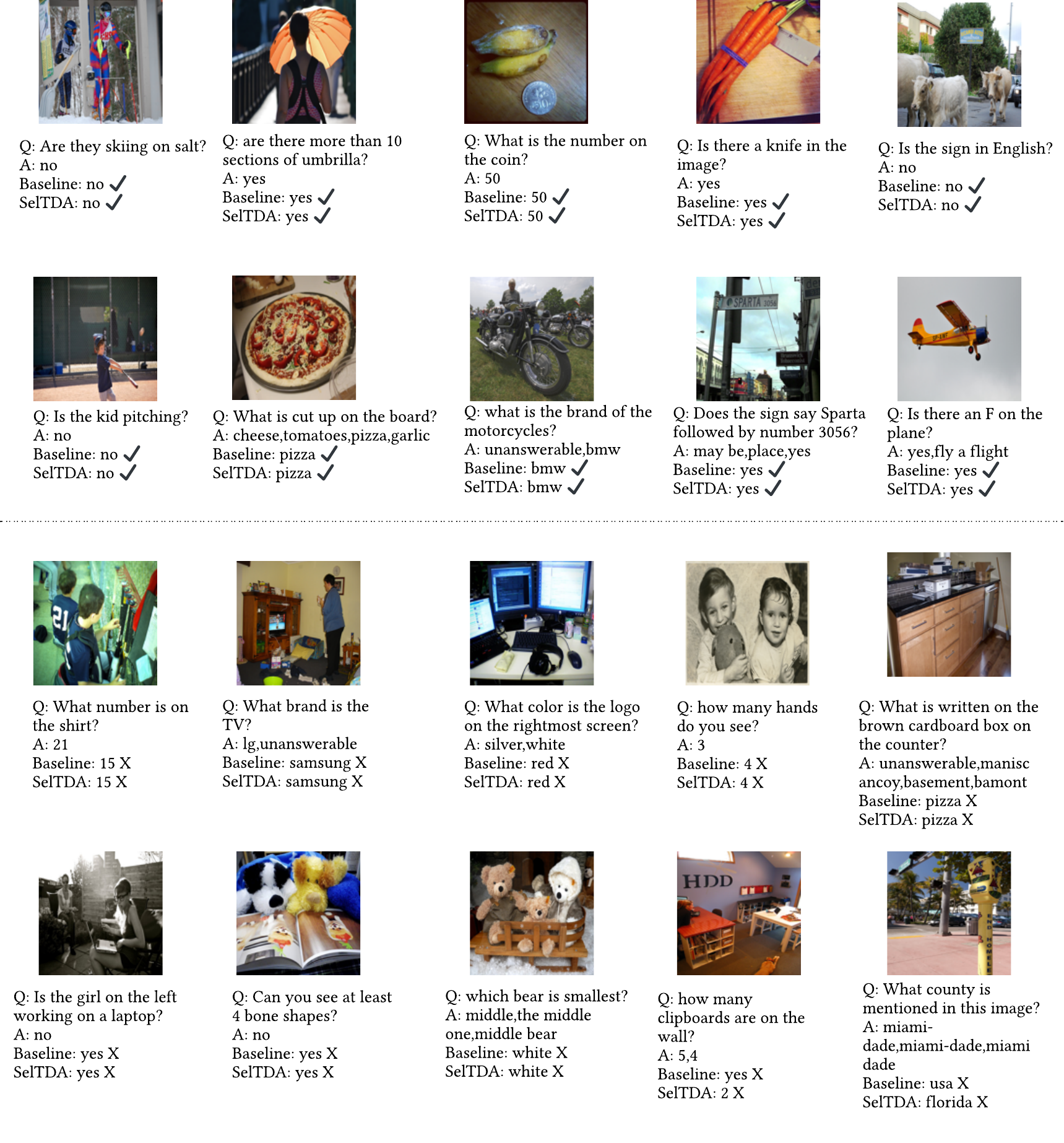}
    \caption{Examples of adversarial questions (AdVQA) that both the baseline model and \stda~trained model get right (top half) and wrong (bottom half).}
    \label{fig:advqa-both-wrong-right}
\end{figure*}

We next turn to questions that both models answer correctly or incorrectly (Fig \ref{fig:advqa-both-wrong-right}). 
Many of the questions that both models answer correctly are boolean questions, or questions that involve reading.
For example, ``Is there an F on the plane?'' or ``What is the number on the coin?''.
Questions that both models answer wrong tend to be questions involving composition of multiple forms of reasoning, such as a referring expression followed by activity recognition, as in ``Is the girl on the left working on a laptop?''.
Interestingly, even in cases where both models are wrong, the model trained with \stda is often \textit{closer} to the correct answer.
For example, in Fig \ref{fig:advqa-both-wrong-right} Row 4, Col 4, the baseline model answers ``yes'' to the question ``How many clipboards are on the wall?'', while the model trained with \stda~answers ``2''. 
The baseline model commits a category error, while the \stda-trained model guesses a number, albeit an incorrect one.
Another example is Fig \ref{fig:advqa-both-wrong-right} Row 4, Col 5, where the baseline model answers ``USA'' to the question ``What \textit{county} is mentioned in this image?'', while the \stda-trained model answers ``Florida''. Both answers are incorrect, as the USA is a \textit{country}, not a county, and Florida is a state rather than a county.
However, the correct answer (Miami-Dade) is a county \textit{within} the state of Florida. 
Although the \stda-trained model is wrong, its wrong answer is more specific and closer to the correct answer than the baseline model.
\section{Domain Generalization Correct / Incorrect Samples}
\begin{figure*}
    \centering
    \includegraphics{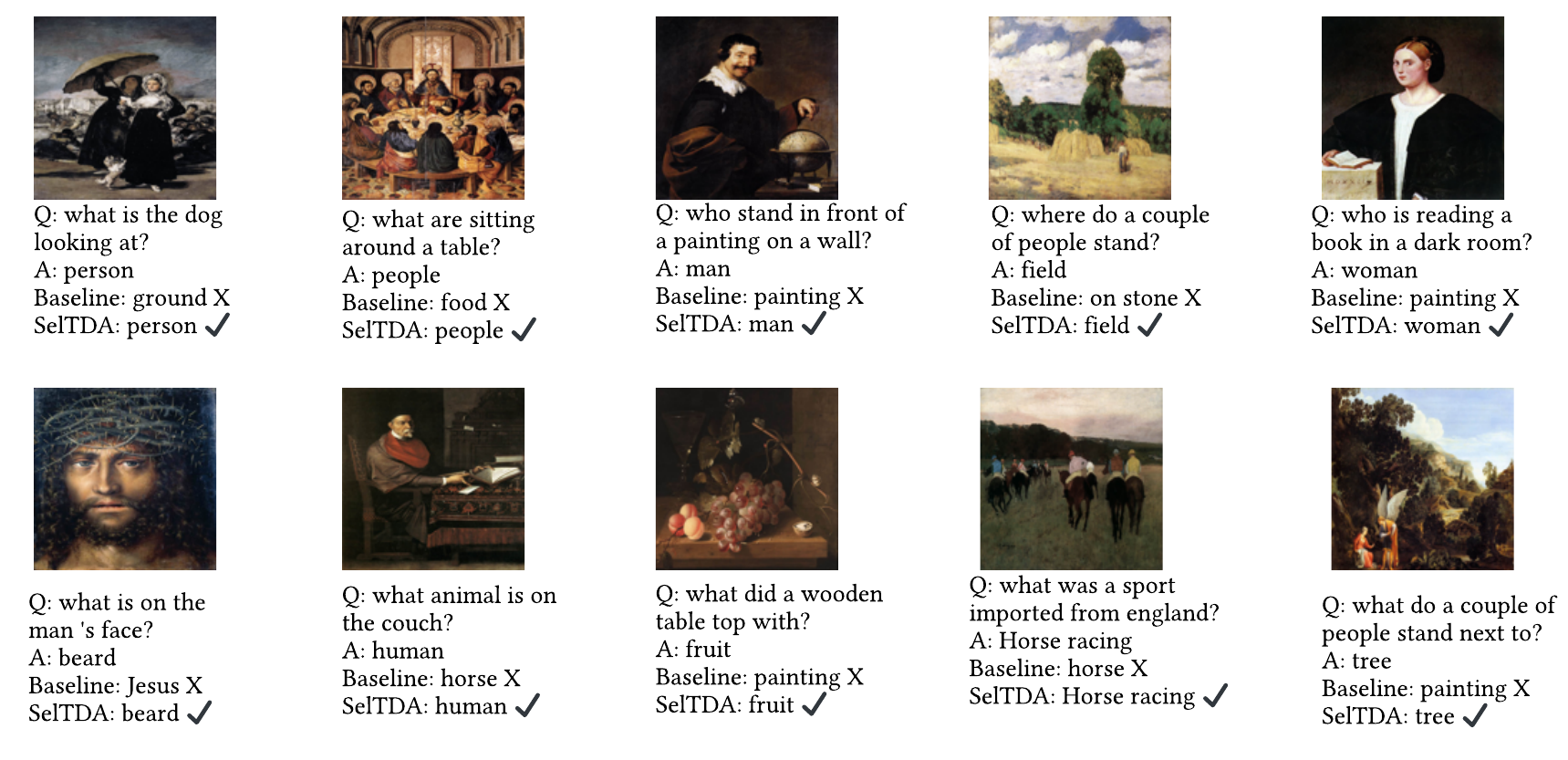}
    \caption{Exampes of fine art questions from AQUA that the baseline model gets wrong, but the model trained with \stda~gets correct.}
    \label{fig:artvqa-right-wrong}
\end{figure*}

We next show results on datasets from different domains, specifically PathVQA (medical) and AQUA (fine art images).
Both the baseline model and the model trained with \stda~were evaluated in a 0-shot manner. 
In Fig \ref{fig:artvqa-right-wrong}, we show questions that the baseline model answers incorrectly, but the model trained with \stda~answers correctly. 
One common failure model of the baseline model is to simply answer ``painting'' to questions, regardless of whether ``painting'' could be a plausible answer to the question.
This suggests that the baseline model is not parsing the question correctly, whereas the model trained with \stda~is able to understand the question.

\begin{figure*}
    \centering
    \includegraphics{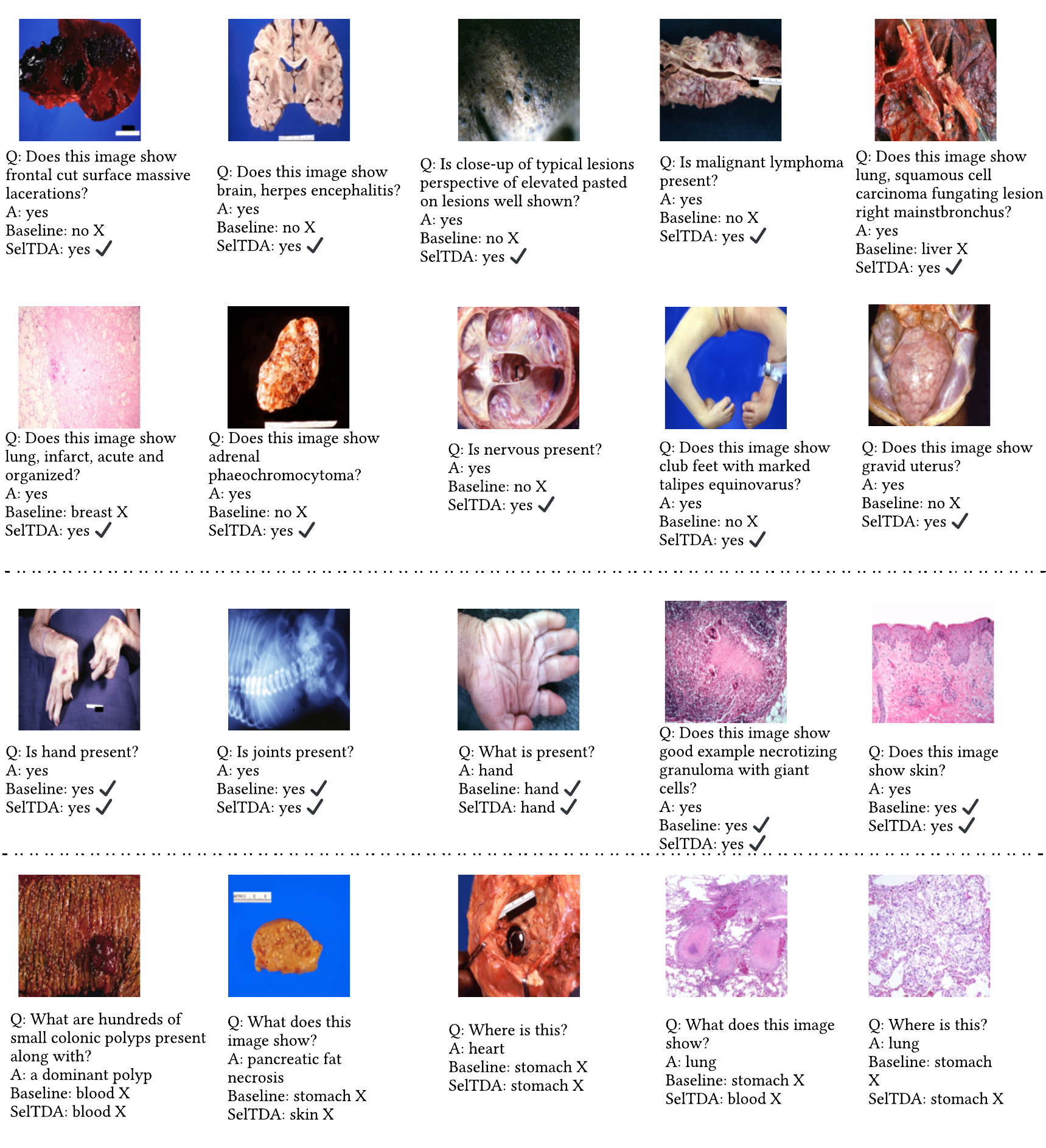}
    \caption{Examples of PathVQA questions that the baseline model gets wrong, but the model trained with \stda~gets correct (top panel), as well as PathVQA questions both models get correct (middle panel) and questions both models get wrong (bottom panel).}
    \label{fig:pathvqa-right-wrong}
\end{figure*}

Next, we show results for PathVQA (Fig \ref{fig:pathvqa-right-wrong}). 
PathVQA is a medical question answering dataset that requires specialized knowledge to answer correctly.
Although the BLIP model was not pretrained on medical data specifically, it was trained on over 100M image-text pairs, and it is likely that some of those image-text pairs included anatomical diagrams or medically relevant information.
Thus, it is reasonable to expect \textit{some} ability to answer questions that require specialized medical knowledge correctly.
We show that this expectation is correct (Fig \ref{fig:pathvqa-right-wrong}, middle panel). 
The questions both models answer correctly tend to be simple identification question such as ``does this image show skin?''.
The model trained with \stda~is able to answer some more challenging questions, such as ``does this image show brain, herpes encephalitis? (yes)'' and ``does this image show club feet with marked talipes equinovarus? (yes)''. 
We hypothesize that the reason the model is able to answer correctly is not necessarily because it is able to identify herpes encephalitis in the brain or identify talipes equinovarus, but because the model can identify more common concepts such as ``club feet'' and ``brain'' and is responding with a ``yes'' based on the presence of those concepts.


\end{document}